\documentclass{article}

\usepackage{arxiv}

\usepackage[utf8]{inputenc} 
\usepackage[T1]{fontenc}    
\usepackage{hyperref}       
\usepackage{url}            
\usepackage{booktabs}       
\usepackage{amsfonts}       
\usepackage{nicefrac}       
\usepackage{microtype}      
\usepackage{xcolor}         
\usepackage{multirow}
\usepackage{caption}
\usepackage{graphicx}
\usepackage{amsmath}
\usepackage{float}
\usepackage{makecell}

\title{EEGChaT: A Transformer-Based Modular Channel Selector for SEEG Analysis}

\author{%
  Chen Wang \\
  School of Computer Science and Technology \\
  University of Science and Technology of China \\
  Hefei, Anhui, China \\
  \texttt{wc2208093719@mail.ustc.edu.cn} \\
  \AND
  Yansen Wang \\
  Microsoft Research Asia \\
  Shanghai, China \\
  \texttt{yansenwang@microsoft.com} \\
  \And
  Dongqi Han \\
  Microsoft Research Asia \\
  Shanghai, China \\
  \texttt{dongqihan@microsoft.com} \\
  \And
  Zilong Wang \\
  Microsoft Research Asia \\
  Shanghai, China \\
  \texttt{wangzilong@microsoft.com} \\
  \And
  Dongsheng Li \\
  Microsoft Research Asia \\
  Shanghai, China \\
  \texttt{Dongsheng.Li@microsoft.com} \\
}

\begin{document}

\maketitle

\begin{abstract}

Analyzing stereoelectroencephalography (SEEG) signals is critical for brain-computer interface (BCI) applications and neuroscience research, yet poses significant challenges due to the large number of input channels and their heterogeneous relevance. 
Traditional channel selection methods struggle to scale or provide meaningful interpretability for SEEG data. 
In this work, we propose EEGChaT, a novel Transformer-based channel selection module designed to automatically identify the most task-relevant channels in SEEG recordings. 
EEGChaT introduces Channel Aggregation Tokens (CATs) to aggregate information across channels, and leverages an improved Attention Rollout technique to compute interpretable, quantitative channel importance scores. 
We evaluate EEGChaT on the DuIN dataset, demonstrating that integrating EEGChaT with existing classification models consistently improves decoding accuracy, achieving up to 17\% absolute gains. 
Furthermore, the channel weights produced by EEGChaT show substantial overlap with manually selected channels, supporting the interpretability of the approach. Our results suggest that EEGChaT is an effective and generalizable solution for channel selection in high-dimensional SEEG analysis, offering both enhanced performance and insights into neural signal relevance.

\end{abstract}

\section{Introduction}
Recent years have witnessed an increasing trend to apply artificial intelligence to understanding physiological signals, especially brain signals such as electroencephalograph (EEG) which attracts great research interests due to its high temporal resolution and relatively low costs\cite{wang2024eegpt,duan2023dewave,li2022multi}. While Scalp EEG is widely used due to its non-invasiveness, it suffers from limited spatial resolution and susceptibility to noise due to the signal being distorted by the skull and scalp. Instead, invasive EEG techniques such as stereoelectroencephalograph (SEEG) demonstrate many merits including its more advantageous signal-to-noise ratio (SNR) and better decodability, thus preferred compared to non-invasive EEG in brain computer interface (BCI), brain health diagnosis, and foundamental neuroscience research\cite{zheng2024discrete,youngerman2019stereoeeg} when the performance is crucial.

While SEEG is effective, its recordings may involve hundreds or even thousands of channels, with both the number of channels and spatial coverage varying greatly across subjects. Such high-dimensional inputs significantly increase the computational burden and make it difficult for models to effectively learn and identify the most informative channels. Moreover, this characteristic of SEEG hinders a generalizable analysis with unified models. In order to get a unified model that can be applied to different SEEG configurations or aggregated analysis among different subjects, it's very important to design a sophisticated method for analyzing channel importance. Ideally, the method should be capable of selecting SEEG channels that are most relevant to the predefined task, in order to reduce computational cost while enhancing decodability.

Previous studies related to channel importance analysis are mostly conducted on scalp EEG, and can be roughly classified as filtering, wrapper, embedded, and human-based techniques\cite{alotaiby2015review}. Nevertheless, the inherent complexity of SEEG signals poses significant challenges for the application of these methods. Filtering methods rely on analyzing the intrinsic characteristics among feature channels\cite{duun2012channel}, while wrapper methods require iterative selection of feature channels based on classification metrics\cite{shih2009sensor}, which is not well-suited for deep learning frameworks. Therefore, filtering and wrapper approaches are generally applicable to EEG data with prominent feature channels or relatively simple classification tasks. In contrast, embedded methods impose higher demands on the channel selection module due to the complexity of the data\cite{strypsteen2021end,wu2024channel}. Currently, such methods are rarely used for SEEG data. Human-based techniques, on the other hand, are typically considered as a last resort, as the functional organization of brain regions is complex and the manual implementation process is time-consuming and labor-intensive\cite{zimbric2011three}.

To this end, we claim that to address the unique challenges posed by SEEG channel selection, it is essential to develop data-driven, model-integrated approaches that can effectively capture the intricate spatial-temporal dynamics and heterogeneous channel relevance inherent in SEEG recordings. We need to design a deep-learning module with the capability to 1) learn channel importance in an end-to-end manner and integrate with the downstream classification or representation learning objectives; 2) incorporate explainable module to represent the channel importance. 3) keep or even boost the general decoding/classification ability of the model. To meet the previous principles, we design EEG Channel Transformer (EEGChaT).
EEGChaT is a Transformer-based\cite{vaswani2017attention} channel selector that treats feature channels as a sequence of tokens. By applying the Attention Rollout technique\cite{abnar2020quantifying}, it traces back the true contribution of each original channel to the Channel Aggregation Token (CAT). These contributions are then used as weights to reweight the original SEEG signals, producing a refined input for the downstream classifier. We conducted several experiments on DuIN dataset\cite{zheng2024discrete} to verify the ability of EEGChat to select channel while keeping an impressive decoding results.


We summarize our contributions as follows.

\textbf{EEGChaT Module:} We propose EEGChaT, a novel channel selection module that automatically identifies the most task-relevant channels in datasets with a large number of channels. Designed as a plug-and-play layer, EEGChaT is compatible with SEEG inputs of arbitrary time durations and channel counts, enabling flexible integration into various neural architectures.\\
\textbf{Improved Performance:} We show that incorporating EEGChaT into downstream classification networks consistently improves performance. For instance, integrating EEGChaT with an EEGConformer backbone on the DuIN dataset boosts classification accuracy by about 13\%, and using EEGChaT with another model yields a 17\% improvement.\\
\textbf{Interpretability:} EEGChaT provides interpretable channel relevance through quantitative weights. The channel weights output by EEGChaT correlate with manually selected channels, offering insight and a degree of trustworthiness in the automated channel-selection process.

\section{Related Works}
\subsection{SEEG Benchmarks}

Due to the lack of high-quality open-source SEEG datasets, we selected only the DuIN dataset as our benchmark. DuIN is a well-annotated Chinese word-reading SEEG dataset, including 12 subjects with pharmacologically intractable epilepsy, which addresses the lack of SEEG language dataset.

Given that we use the DuIN dataset, we selected the two most representative classification networks from the original paper: the Du-IN Encoder and EEGConformer\cite{song2022eeg}. The Du-IN Encoder, which integrates Vector Quantized Variational Autoencoder (VQ-VAE)\cite{van2017neural} and Masked Autoencoder (MAE)\cite{he2022masked}, achieved state-of-the-art (SOTA) performance using manually selected channels.

\subsection{Channel Importance Analysis}

However, there are a very limited number of works related to channel importance analysis on SEEG data. The first study\cite{strypsteen2021end} employed Concrete Autoencoders as the channel selector, leveraging the Gumbel-Softmax trick\cite{balin2019concrete} for discrete sampling while enabling continuous gradient flow. The second study\cite{wu2024channel} adopted Stochastic Gates (STG)\cite{yamada2020feature} as the channel selector, utilizing multiple Bernoulli gates for discrete sampling and applying Gaussian reparameterization to ensure proper gradient propagation.

However, both methods have several limitations. First, they suffer from insufficient learning capacity and are unable to provide interpretable channel weights. Additionally, the Gumbel-Softmax method requires the number of output channels to be predefined and may select the same channel multiple times. Meanwhile, the STG method faces difficulties in accurately estimating the true number of activated channels.
\section{Methodology}
In this section, we formally present the definition of the channel importance analysis task on SEEG, and detail our design of EEGChaT.

\subsection{Task Definition}

SEEG datasets are typically represented as a specific type of time-series dataset $\mathcal{D} = \{(\mathbf{x}_i, y_i)\}_{i=1}^M$, where the SEEG segment $\mathbf{x}_i \in \mathbb{R}^{C_i \times L}$ corresponds to a specific brain state $y_i \in \{0, 1, ..., S-1\}$. Here, $C_i = |\mathcal{C}_i|$ denotes the number of channels for the $i$-th sample, which may vary across different subjects due to individualized electrode implantation schemes. $L = f_s \times T$ denotes the number of time steps, determined by the sampling rate $f_s$ and the segment duration $T$. 

Unlike a purely classification-based task, the channel importance analysis task aims to learn a model $ f_{\theta} $ parameterized by $ \theta $. On one hand, the output $ \mathbf{x}'_i $ of $ f_{\theta} $ is used as the input to a downstream classifier $ f_{\gamma} $, parameterized by $ \gamma $, to predict the brain state $ \hat{y}_i $. On the other hand, $ f_{\theta} $ internally produces an intermediate result $ \mathbf{W} \in \mathbb{R}^{N \times C} $, where $ N $ is determined by the initialization of $ f_{\theta} $. This intermediate result serves as a weight matrix and is multiplied with the input to generate the output $ \mathbf{x}'_i=\mathbf{W}\mathbf{x_i} \in \mathbb{R}^{N \times L} $ of $ f_{\theta} $.

\subsection{EEGChaT Framework}

\begin{figure}[htbp] 
  \centering
  \includegraphics[width=0.8\textwidth]{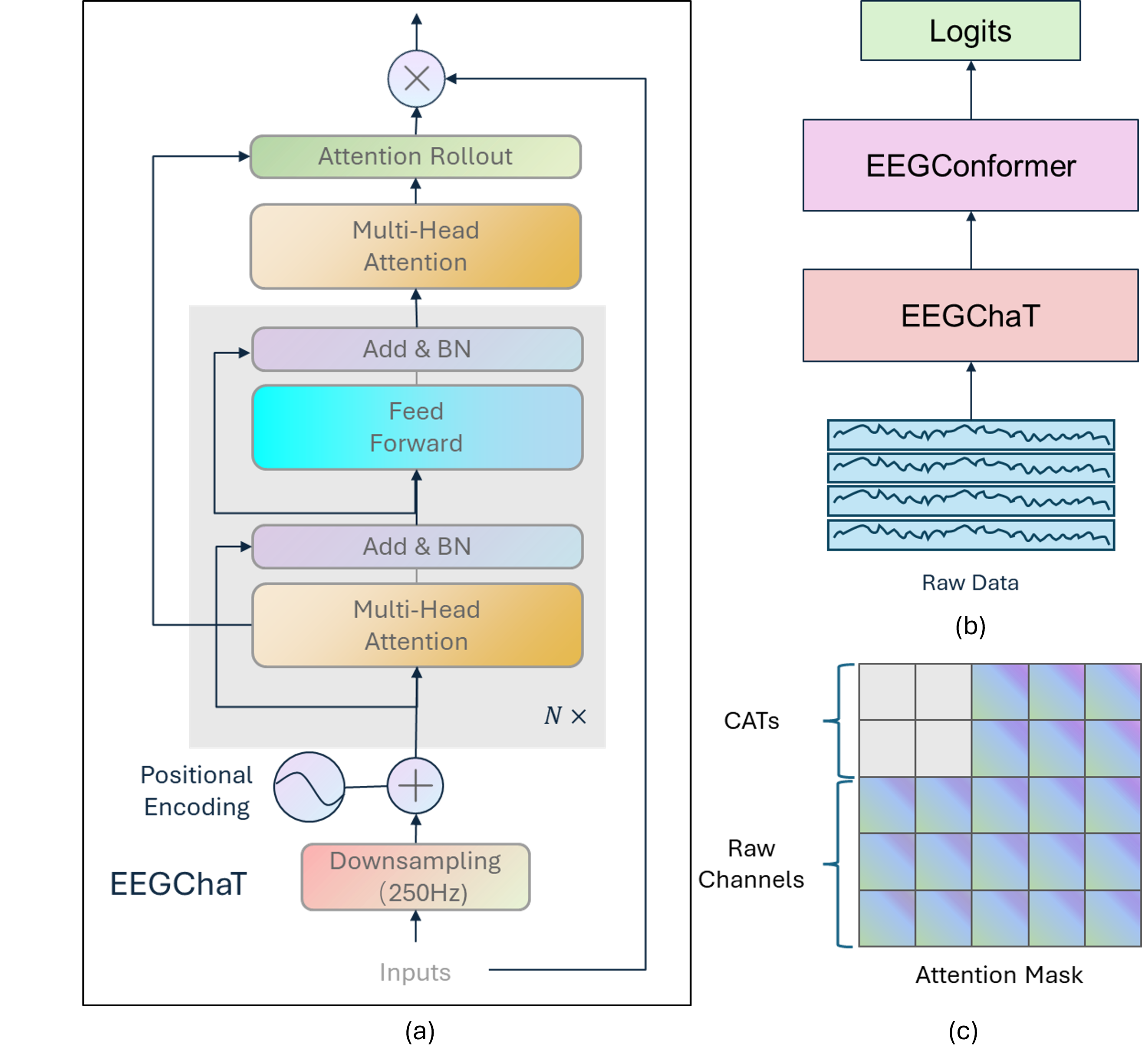}
  \caption{The framework of EEGChaT. (a) Overview of our methods. The input data to EEGChaT is downsampled to 250Hz to reduce the complexity of channel feature extraction for the model. (b) EEGChaT and downstream classifier integration diagram. (c) Default attention mask strategy in EEGChaT's multi-head self-attention. The gray areas in the figure indicate regions where attention score computation is not performed.}
  \label{fig:EEGChaT}
\end{figure}

We introduce EEGChaT, a Transformer-based plug-and-play channel selector that explicitly outputs normalized importance weights for raw channels. EEGChaT achieves this functionality through the following two components: 1) Transformer layer with the Channel Aggregation Token(CAT). By transposing the original input, SEEG signals are fed into a Transformer encoder layer where each token corresponds to a raw channel. A specially designed attention masking mechanism enables the CAT to aggregate information from all channels. 2) Importance analysis with Attention Rollout. Inspired by the Attention Rollout technique, we trace back the contribution of each raw channel to the CAT across multiple layers of attention matrices, ultimately deriving the normalized importance weights for the original channels.

\subsubsection{Transformer Layer with the Channel Aggregation Token}

Transformer models have become widely adopted for processing time series data. For time series classification and related tasks, time points or time patches (i.e., continuous segments of time points aggregated through linear or nonlinear transformations) are typically used as tokens. To enable CAT to selectively aggregate information from the original channels using attention matrices, an intuitive approach is to treat the original channels as tokens, with the temporal dimension serving as the hidden state.

SEEG data typically have a high sampling rate. To reduce the computational cost and facilitate channel-specific feature extraction, we downsample the input signals to 250 Hz before feeding them into EEGChaT:
\begin{equation}
    \textbf{z}_j=\text{Downsample}(\textbf{x}_{i,j})\in\mathbb{R}^d
\end{equation}
where $\textbf{x}_{i,j} \in \mathbb{R}^L$ denotes the raw signal from the $j$-th channel, and the downsampled vector $\textbf{z}_j$ is subsequently treated as a channel token.

Then, we stack all channel embeddings and prepend $n_{cat}$ learnable Channel Aggregation Tokens (CATs), denoted as $\textbf{z}^{CAT}i \in \mathbb{R}^d$ ($i=1,\dots,n_{cat}$), to the sequence:
\begin{equation}
    \textbf{Z}=[\textbf{z}_1;\textbf{z}_2;\dots;\textbf{z}_C]\in\mathbb{R}^{C\times d}
\end{equation}
\begin{equation}
    \textbf{Z}'=[\textbf{z}^{CAT}_1;\dots;\textbf{z}^{CAT}_{n_{cat}};\textbf{z}_1;\dots;\textbf{z}_C]\in\mathbb{R}^{(n_{cat}+C)\times d}
\end{equation}
where $n_{cat}$ denotes the number of the CATs.

After incorporating the learnable positional encoding, the channel sequence is fed into the Transformer encoder layer. Compared to the conventional Transformer encoder architecture, we introduced two key modifications. First, we designed a customized masking mechanism to ensure that different CATs can learn distinct importance patterns from the original channels:
\begin{equation}
\text{Attention}(Q, K, V, M) = \text{Softmax}\left( \frac{QK^\top}{\sqrt{d_k}} + M \right)V
\end{equation}
where $M_{ij} = -\infty$ for masked positions and $0$ otherwise. Our default masking strategy, as illustrated in Fig.~\ref{fig:EEGChaT}(c), is designed such that no information is exchanged between different CATs.

Second, we employed Batch Normalization to stabilize the training process:
\begin{equation}
\text{BatchNorm}(x + \text{Sublayer}(x))
\end{equation}

The overall architecture design is illustrated in Fig.~\ref{fig:EEGChaT}(a).

\subsubsection{Importance Analysis with Attention Rollout}

\begin{figure}[htbp] 
  \centering
  \includegraphics[width=\linewidth]{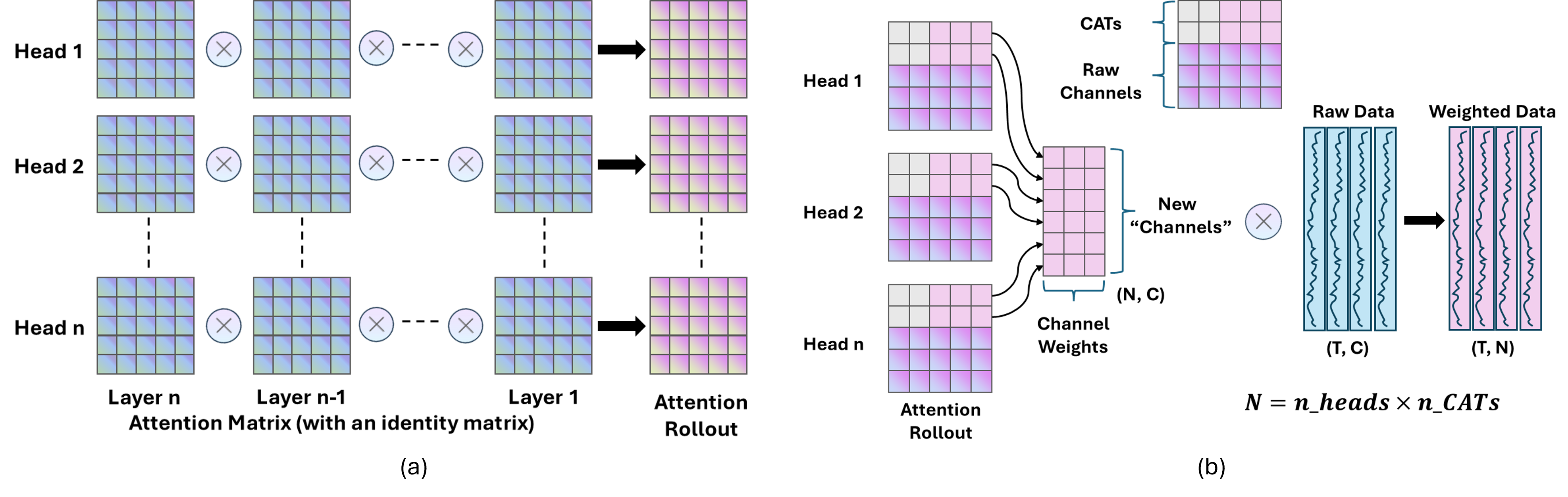}
  \caption{Deriving channel weights from attention matrices. (a)Tracing back attention scores from raw channels to the CAT using Attention Rollout across multiple layers. (b) Deriving the weight tensor of raw channels via the Attention Rollout.}
  \label{fig:weights}
\end{figure}

In the Transformer model, each layer's self-attention mechanism generates distinct attention scores, which reflect the model's focus on different input features at various depths. The attention distributions in each layer carry unique significance—lower layers typically attend to local and detailed features, while higher layers tend to capture more global and abstract representations\cite{tenney2019bert,yuan2021tokens}.

Attention Rollout\cite{abnar2020quantifying} integrates information across layers by accumulating the attention scores from each layer, thereby forming a global attention map. In the original Attention Rollout method, the attention scores from different heads within the same layer are averaged:
\begin{equation}
Avg_h\left(A^{(l)}\right) = \frac{1}{H} \sum_h A_h^{(l)}
\label{eq:Eattention}
\end{equation}
Here, $A^{(l)}$ denotes the attention score matrix at layer $l$, obtained by adding the identity matrix to the attention map to account for residual connections, and $H$ represents the number of attention heads. This averaging approach assumes that all heads are equally important. However, in practice, different attention heads often capture distinct features, and the importance of these features can vary significantly\cite{clark2019does}.

Based on this, we hypothesize that attention heads at the same position across different layers can learn temporally coherent features under the mechanism’s constraints, with increasing depth abstracting local features into higher-level representations. Therefore, when computing Attention Rollout, we retain the attention head dimension, resulting in an Attention Rollout for each attention head. The calculation process is illustrated in Fig.~\ref{fig:weights}(a). In the figure, each matrix operation is preceded by row-wise normalization of the attention matrix.

Let the resulting Attention Rollout be denoted as $\mathcal{R}\in\mathbb{R}^{H\times(n_{cat}+C)\times(n_{cat}+C)}$. It still requires a series of transformations to obtain the final matrix $\textbf{W}\in\mathbb{R}^{C\times(H*n_{cat})}$. The transformation process is illustrated in Fig.~\ref{fig:weights}(b). 

Since, for each CAT, we only extract its attention scores with respect to the original channels, a final row-wise normalization is still required for $\textbf{W}$.

Finally, we perform a matrix multiplication between the obtained channel importance weights $\textbf{W}$ and the original input $\mathbf{x_i}$ (i.e., the SEEG signals before downsampling), resulting in the input to the downstream classifier filtered by EEGChaT. Subsequently, the result can be fed into the downstream classifier as shown in Fig.~\ref{fig:EEGChaT}(b).


\section{Experiments}
In this section, we conduct experiments to answer the following research questions (RQ):
\begin{itemize}
    \item \textbf{RQ1}: Can EEGChaT still achieve competitive performance compared to baselines?
    \item \textbf{RQ2}: Are the tokens selected by EEGChaT really important for the downstream tasks?
\end{itemize}

\subsection{Datasets and Baselines}

\paragraph{DuIN\cite{zheng2024discrete}.} Due to the lack of high-quality open-access SEEG datasets, we conduct our experiments exclusively on the DuIN dataset. DuIN is a well-annotated Chinese word-reading sEEG dataset (vocal production), including 12 subjects. The subjects undergo a surgical procedure to implant 7 to 13 invasive sEEG electrodes, each with 72 to 158 channels, in their brain. For each subject, the dataset contains 15 hours of 2000Hz recordings, 3 hours of which are task recordings. In the task, subjects were instructed to watch and read aloud randomly presented Chinese words on the screen. These words were preselected based on specific criteria, totaling 61 in number. Since only the task recordings were released by the authors, we adopt supervised training as our primary training strategy to ensure the reproducibility of the results.

To verify the performance of EEGChat, we compare our model with the following baselines:
\paragraph{EEGConformer\cite{song2022eeg}.} EEGconformer is a compact convolutional transformer, which can encapsulate local and global features in a unified EEG classification framework. Specifically, the convolution module learns the low-level local features throughout the one-dimensional temporal and spatial convolution layers. The self-attention module is straightforwardly connected to extract the global correlation within the local temporal features. Subsequently, the simple classifier module based on fully-connected layers is followed to predict the categories for EEG signals.
\paragraph{Du-IN Encoder\cite{zheng2024discrete}.} Du-IN Encoder is a general architecture for SEEG speech decoding tasks that can deal with any input SEEG signals with arbitrary time length. The key operation for archiving this is segmenting the SEEG signals into patches. The Du-IN Encoder consists of three main components: the Spatial Encoder, Temporal Embedding, and Transformer Encoder.

For both downstream classifiers, we adopt the default parameters provided in the DuIN open-source repository, which are specifically configured for the DuIN dataset.

\subsection{Implementation details}
For fair comparison, we keep the hyperparameters the same across all the baselines as much as possible. 
\paragraph{Preprocess.} We closely followed the data preprocessing procedure described in the original DuIN paper. Each subject's 3000 samples were split into training, validation, and test sets in an $8{:}1{:}1$ ratio. We adopted the same data augmentation strategies as used in the original study. However, there is one notable difference: we applied Z-score normalization on a per-sample, per-channel basis. This approach has been shown to better facilitate EEGChaT in focusing on high-level features of the original channels rather than their statistical means or variances.
\paragraph{Training.} We used the same training parameters for both baseline models. We employed the AdamW optimizer with a weight decay of $5 \times 10^{-3}$ and a batch size of 32. The max learning rate was set to $2 \times 10^{-4}$ ($1 \times 10^{-4}$ when integrated with EEGChaT, which can improve the stability of model training). The learning rate scheduler adopted the OneCycleLR strategy, with a warm-up ratio set to 0.3. A cosine annealing strategy was employed, and the final learning rate decayed to one-tenth of the maximum learning rate. The maximum number of epochs was set to 200. All experiments were repeated with 9 different random seeds to reduce stochastic variation.

\subsection{Results for Downstream Task Prediction}
In this section, we apply the above two baseline models on the DuIN dataset with the full set of channels. The prediction results are shown in Table~\ref{tab:all_chans}. For clarity and aesthetics, EEGConformer is abbreviated as \textbf{CFMR}.

\begin{table}[ht]
\centering
\caption{Performance of different methods across subjects (best in \textbf{bold}, second best \underline{underlined}). *: EEGChaT improved both models' performance for this subject. **: EEGChaT outperformed all models without EEGChaT for this subject.}
\label{tab:all_chans}
\begin{tabular}{lcccc}
\toprule
\multirow{2}{*}{Subject} & \multicolumn{4}{c}{Accuracy $\pm$ StdDev (\%)} \\
\cmidrule(lr){2-5}
& CFMR & EEGChaT+CFMR & Du-IN & EEGChaT+DuIN \\
\midrule
$1^{**}$  & 47.56 $\pm$ 6.26 & \underline{60.26 $\pm$ 4.44} & 57.65 $\pm$ 5.06 & \textbf{67.61 $\pm$ 1.85} \\
$2^{**}$  & 39.91 $\pm$ 5.19 & \underline{58.62 $\pm$ 5.77} & 49.34 $\pm$ 4.77 & \textbf{70.00 $\pm$ 2.01} \\
$3$  & \textbf{8.86 $\pm$ 1.50}  & \underline{8.86 $\pm$ 2.89}  & 4.48 $\pm$ 1.97  & 8.53 $\pm$ 2.76 \\
$4^{**}$  & 13.73 $\pm$ 1.69 & \underline{44.59 $\pm$ 4.29} & 15.70 $\pm$ 3.07 & \textbf{53.77 $\pm$ 4.40} \\
$5^{**}$  & 31.37 $\pm$ 4.99 & \underline{58.94 $\pm$ 5.16} & 38.40 $\pm$ 6.87 & \textbf{71.51 $\pm$ 5.11} \\
$6^{**}$  & 6.50 $\pm$ 0.88  & \underline{15.69 $\pm$ 4.48} & 5.21 $\pm$ 2.37  & \textbf{25.16 $\pm$ 5.46} \\
$7^{**}$  & 17.68 $\pm$ 2.88 & \underline{30.02 $\pm$ 8.98} & 16.87 $\pm$ 2.78 & \textbf{39.69 $\pm$ 3.04} \\
$8$  & \underline{12.15 $\pm$ 1.77} & 11.85 $\pm$ 2.97 & 9.37 $\pm$ 3.93  & \textbf{13.25 $\pm$ 5.27} \\
$9^{**}$  & 31.51 $\pm$ 3.86 & \underline{56.19 $\pm$ 4.03} & 45.88 $\pm$ 6.15 & \textbf{66.41 $\pm$ 3.27} \\
$10^{**}$ & 6.84 $\pm$ 1.96  & \textbf{13.39 $\pm$ 2.95} & 6.81 $\pm$ 1.46  & \underline{10.70 $\pm$ 4.65} \\
$11^{**}$ & 14.02 $\pm$ 2.48 & \underline{26.64 $\pm$ 12.27} & 20.99 $\pm$ 7.71 & \textbf{39.43 $\pm$ 12.82} \\
$12^*$ & 8.26 $\pm$ 2.85  & 8.80 $\pm$ 3.73  & \underline{9.11 $\pm$ 2.96}  & \textbf{16.58 $\pm$ 5.65} \\
\bottomrule
Avg &19.87        &   \underline{32.82}      &     23.32    &   \textbf{40.22}  \\
\bottomrule
\end{tabular}
\end{table}

From a holistic perspective, when it is unclear which SEEG channels are more relevant to the downstream task, EEGChaT can enhance model performance within a short time without the need for meticulous channel selection. Moreover, the accuracy improvement is more pronounced for models with stronger baseline performance (e.g., the Du-IN Encoder saw an improvement of approximately 17\%, while EEGConformer improved by only around 13\%).

At the individual level, EEGChaT demonstrates a certain degree of robustness across subjects, with nearly all participants showing substantial increases in classification accuracy after incorporating the EEGChaT module. Here, we answer the \textbf{RQ1}.

However, it should also be noted that for some subjects, the use of EEGChaT for channel selection may lead to instability in training outcomes. This phenomenon is also related to the performance of the downstream classifier. Baseline models with stronger learning capabilities can better assist EEGChaT in identifying task-relevant channels, thereby reducing performance fluctuations caused by different random seeds.

\begin{figure}[htbp] 
  \centering
  \includegraphics[width=\linewidth]{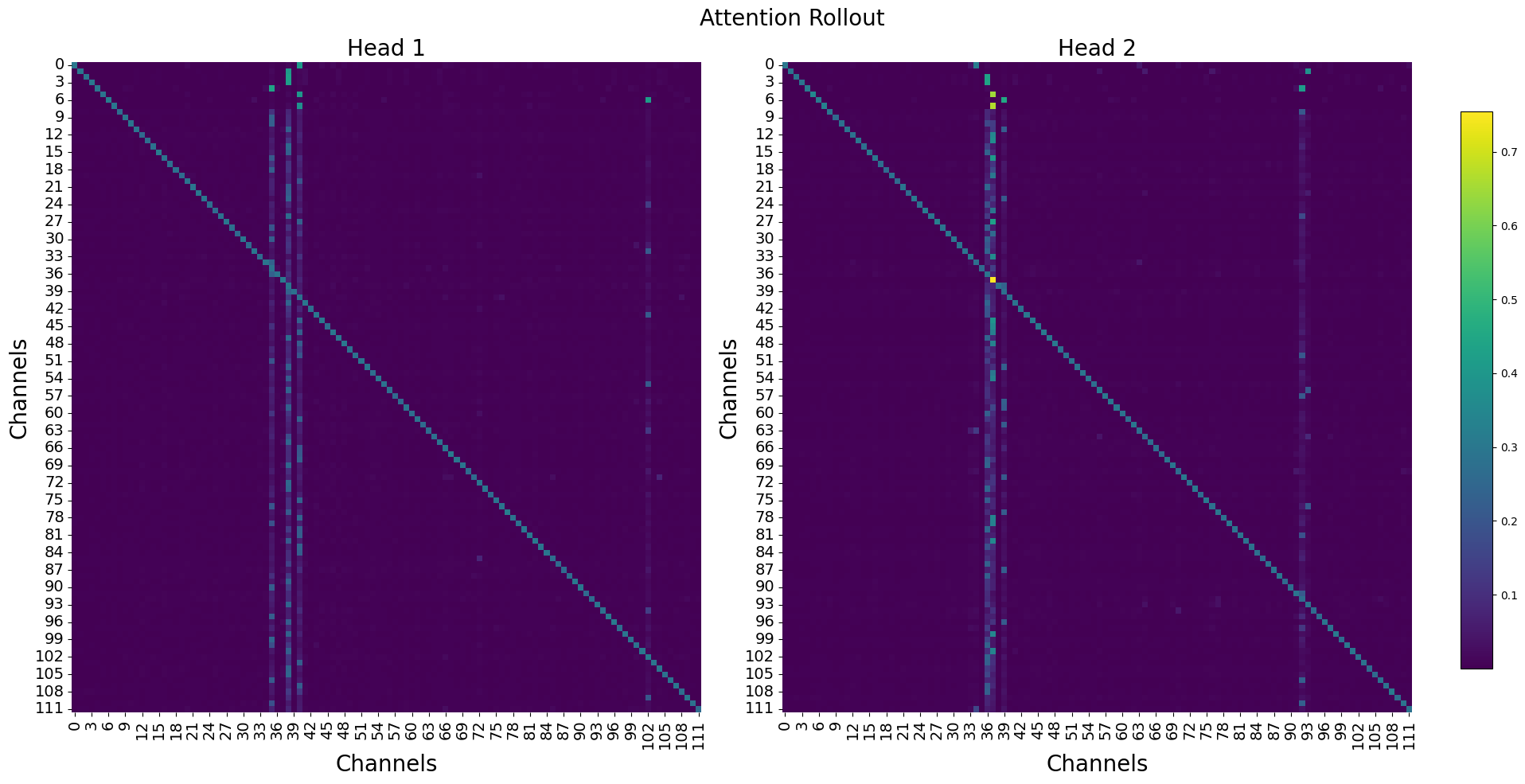}
  \caption{The averaged Attention Rollout of Subject 1 on the test set.}
  \label{fig:AttentionRollout}
\end{figure}

In addition, we visualized the averaged Attention Rollout of Subject 1 across all test set samples, as shown in Fig.~\ref{fig:AttentionRollout}. In our experiments, the number of CATs was set to 8 and the number of attention heads to 2. Therefore, channels with indices 0 to 7 correspond to the CATs, while channels starting from index 8 correspond to the original channels, beginning with index 0.

From the figure, we can validate our previous assumptions:
\begin{itemize}
    \item Different CATs are capable of capturing different task-relevant channels.
    \item Due to the constraints imposed by the Attention Rollout computation, attention heads with the same index across different layers tend to focus on similar task-relevant channels.
\end{itemize}
Additionally, the figure reveals that task-relevant SEEG channels exhibit sparsity, which is consistent with prior medical knowledge.




\subsection{Results for Important Channel Analysis}
We also verified the importance of channels selected by our model by run models on the selected channels. Results can be found in Table~\ref{tab:sel_chans_EEGConformer}. For clarity and aesthetics, \textbf{EC} refers to the channels selected by EEGChaT, while \textbf{MC} denotes the manually selected channels. All results are obtained based on the EEGConformer model.

\begin{table}[htbp]
\centering
\caption{Performance of different methods across subjects (best in \textbf{bold}, second best \underline{underlined}). The Jaccard Index column is computed based on the overlap between the 10 hand-crafted channels and the 10 highest-weighted channels selected by EEGChaT (obtained by averaging the weights across all samples, attention heads, CATs, and random seeds).}
\label{tab:sel_chans_EEGConformer}
\resizebox{1\textwidth}{!}{
\begin{tabular}{cccccc}
\toprule
\multirow{2}{*}{Subject} & \multicolumn{4}{c}{Accuracy $\pm$ StdDev (\%)} & \multirow{2}{*}{\makecell{Jaccard\\ Index}} \\
\cmidrule(lr){2-5}
& EC w/ EEGChaT & EC w/o EEGChaT & MC w/ EEGChaT & MC w/o EEGChaT & \\
\midrule
1  & \textbf{59.83 $\pm$ 4.07} & 55.66 $\pm$ 3.79 & \underline{59.69 $\pm$ 2.42} & 58.92 $\pm$ 3.60 & 0.6667 \\
2  & \underline{62.84 $\pm$ 2.28} & 58.49 $\pm$ 3.88 & \textbf{64.55 $\pm$ 4.07} & 60.36 $\pm$ 2.94 & 0.5385 \\
3  & 11.92 $\pm$ 1.80 & 9.08 $\pm$ 0.98  & \textbf{17.84 $\pm$ 2.57} & \underline{16.65 $\pm$ 1.28} & 0.4286 \\
4  & \textbf{55.08 $\pm$ 1.48} & 46.92 $\pm$ 4.87 & \underline{53.48 $\pm$ 4.60} & 42.40 $\pm$ 3.56 & 0.6667 \\
5  & 64.26 $\pm$ 1.67 & 56.76 $\pm$ 2.70 & \textbf{66.52 $\pm$ 5.59} & \underline{64.66 $\pm$ 2.76} & 0.5385 \\
6  & 22.49 $\pm$ 4.59 & 21.56 $\pm$ 2.26 & \textbf{23.28 $\pm$ 3.76} & \underline{23.08 $\pm$ 2.99} & 0.4286 \\
7  & \underline{38.22 $\pm$ 3.68} & 32.58 $\pm$ 3.77 & \textbf{40.67 $\pm$ 2.15} & 34.90 $\pm$ 3.17 & 0.6667 \\
8  & 25.93 $\pm$ 4.60 & 17.85 $\pm$ 1.56 & \textbf{35.60 $\pm$ 1.75} & \underline{31.96 $\pm$ 1.85} & 0.4286 \\
9  & \underline{56.00 $\pm$ 3.42} & 48.31 $\pm$ 3.40 & \textbf{56.96 $\pm$ 2.62} & 51.98 $\pm$ 2.79 & 0.4286 \\
10 & \underline{17.51 $\pm$ 1.27} & 15.19 $\pm$ 1.77 & \textbf{18.30 $\pm$ 1.47} & 17.06 $\pm$ 1.87 & 0.5385 \\
11 & 42.20 $\pm$ 4.19 & 31.23 $\pm$ 2.65 & \textbf{54.50 $\pm$ 6.68} & \underline{45.92 $\pm$ 2.99} & 0.6667 \\
12 & 13.84 $\pm$ 4.36 & 13.72 $\pm$ 2.18 & \underline{25.32 $\pm$ 2.22} & \textbf{28.35 $\pm$ 3.86} & 0.3333 \\
\bottomrule
Avg & \underline{39.18} & 33.95 & \textbf{43.06} & 39.69 & 0.5275 \\
\bottomrule
\end{tabular}
}
\end{table}

From the table, we can draw the following conclusions:
\begin{itemize}
    \item Constraining the channel selection to a narrower predefined range helps improve the training stability of EEGChaT. This suggests that when using EEGChaT to assist in identifying task-relevant channels, it is beneficial to first narrow down a rough region based on medical prior knowledge.
    \item From the Jaccard Index perspective, the channels selected by EEGChaT generally show a high degree of overlap with those selected manually, and the model performance is also comparable. This addresses \textbf{RQ2} and indicates that our method exhibits a certain level of interpretability.
    \item By comparing the results of "EC w/ EEGChaT" with the previously reported EEGChaT+CFMR results, it can be seen that EEGChaT possesses the ability to further refine and purify the set of channels it initially selects.
\end{itemize}


\section{Limitations}

Although our method has been validated on the DuIN dataset, further evaluation on additional datasets and baseline models is necessary to demonstrate the generalizability of EEGChaT. However, this effort is constrained by the limited availability of high-quality open-source datasets and related prior work. In addition, EEGChaT requires more robust mechanisms to enhance its training stability.
\section{Conclusion}

In this paper, we introduce EEGChaT, a channel selector tailored for SEEG data with a large number of channels. EEGChaT computes channel-wise importance using a Transformer equipped with Channel Attention Tokens (CATs), and enhances interpretability through the use of the Attention Rollout technique. We validate the effectiveness of EEGChaT based on the Du-IN framework, demonstrating that its performance is comparable to that of manually selected channels. On one hand, EEGChaT provides a benchmark for channel selection in SEEG and more broadly in invasive EEG research; on the other hand, our results offer empirical support for the statement that "attention is not not explanation\cite{wiegreffe2019attention}." We hope our work can inspire future research on model interpretability based on Transformer architectures.

\section*{Broader Impact}
\label{sec:broader_impact}
This work aims to advance the field of SEEG analysis.
We do not see any negative societal impacts of this work.

\bibliographystyle{unsrt}
\bibliography{main}

\newpage
\appendix

\section{Hyperparameter Study}
\label{sec:appendix_A}

To evaluate the impact of architectural choices on the performance of our model, we systematically varied several key parameters, including the number of Transformer layers, attention heads, Channel Aggregation Tokens (CATs), and the dropout probability in the Transformer. All experimental settings follow the implementation details described in Section 4.2 of the main text, except for the architectural hyperparameters of EEGChaT, which are varied in this study. We used EEGConformer as the downstream classifier.The initial values of the network parameters under investigation are as follows:
\begin{itemize}
    \item Number of \textbf{Transformer layers}: 2
    \item Number of \textbf{attention heads}: 2
    \item Number of \textbf{CATs}: 8
    \item \textbf{Dropout probability} in the Transformer: 0.3
\end{itemize}
These settings correspond to the network parameters used for EEGChaT in the main experiments. Unless otherwise specified, the EEGChaT parameters used in the appendix are the same as those described above.

\subsection{Number of Transformer Layers}

\paragraph{Settings.} We set the number of Transformer layers to 2, 4, 6, and 8, while keeping the other network parameters fixed for the study.

\begin{table}[htbp]
\centering
\caption{Performance of different network parameter settings across subjects (best in \textbf{bold}, second best \underline{underlined}).}
\label{tab:layers}
\begin{tabular}{c c c c c}
\toprule
\multirow{2}{*}{Subject} & \multicolumn{3}{c}{Accuracy $\pm$ Std (\%)} \\
\cmidrule(lr){2-4}
 & 2 & 4 & 6 \\
\midrule
1  & \textbf{59.75 $\pm$ 4.62} & \underline{59.10 $\pm$ 3.00} & 44.63 $\pm$ 11.78 \\
2  & \textbf{58.31 $\pm$ 5.61} & \underline{51.79 $\pm$ 9.54} & 30.43 $\pm$ 9.16 \\
3  & \textbf{8.92 $\pm$ 2.82}  & \underline{7.28 $\pm$ 2.50}  & 3.89 $\pm$ 2.38  \\
4  & \textbf{44.92 $\pm$ 3.68} & \underline{39.56 $\pm$ 7.40} & 28.01 $\pm$ 12.80 \\
5  & \textbf{58.91 $\pm$ 5.65} & \underline{55.01 $\pm$ 9.57} & 50.49 $\pm$ 7.36 \\
6  & \textbf{16.06 $\pm$ 4.96} & \underline{13.35 $\pm$ 3.46} & 8.74 $\pm$ 3.31  \\
7  & \textbf{29.73 $\pm$ 9.00} & \underline{25.88 $\pm$ 8.53} & 11.08 $\pm$ 7.40 \\
8  & \textbf{11.85 $\pm$ 2.97} & \underline{10.65 $\pm$ 5.68} & 5.54 $\pm$ 2.51  \\
9  & \textbf{56.19 $\pm$ 4.03} & \underline{53.29 $\pm$ 6.19} & 43.87 $\pm$ 8.57 \\
10 & \textbf{13.42 $\pm$ 2.99} & \underline{10.16 $\pm$ 3.42} & 5.59 $\pm$ 1.83  \\
11 & \textbf{26.85 $\pm$ 12.07}& \underline{8.87 $\pm$ 6.62}  & 6.76 $\pm$ 2.39  \\
12 & \underline{8.32 $\pm$ 3.54}  & \textbf{8.93 $\pm$ 4.05}  & 7.77 $\pm$ 2.53  \\
\bottomrule
Avg & \textbf{32.77}              & \underline{28.66}         & 20.57    \\
\bottomrule
\end{tabular}
\end{table}

\paragraph{Observation.} The experimental results are shown in Table~\ref{tab:layers}. Note that the results for the configuration with 8 Transformer layers are not included, as most of the corresponding experiments failed to converge. This suggests that, as the number of layers increases, the potential improvement in the model’s ability to capture channel-wise importance is outweighed by the increased difficulty in training deeper networks. The results from the remaining configurations also show a general decline in model performance as the number of layers increases. However, for subject 12, a slight performance improvement is observed when the number of layers is set to 4. This indicates that increasing the depth of the network can enhance its ability to extract task-relevant channels in more complex channel scenarios.

\subsection{Number of Attention Heads}

\paragraph{Settings.} We set the number of attention heads to 2, 3, 5, and 10, while keeping the other network parameters fixed for the study.

\begin{table}[htbp]
\centering
\caption{Performance of different network parameter settings across subjects (best in \textbf{bold}, second best \underline{underlined}).}
\label{tab:heads}
\begin{tabular}{c c c c c c}
\toprule
\multirow{2}{*}{Subject} & \multicolumn{4}{c}{Accuracy $\pm$ Std (\%)} \\
\cmidrule(lr){2-5}
 & 2 & 3 & 5 & 10 \\
\midrule
1  & 60.26 $\pm$ 4.44 & \underline{63.37 $\pm$ 3.21} & 63.34 $\pm$ 4.56 & \textbf{65.15 $\pm$ 2.55} \\
2  & 58.62 $\pm$ 5.77 & \underline{63.54 $\pm$ 3.72} & 62.41 $\pm$ 5.40 & \textbf{66.02 $\pm$ 2.23} \\
3  & 8.86 $\pm$ 2.89  & \underline{11.95 $\pm$ 4.09} & 9.24 $\pm$ 1.91  & \textbf{12.98 $\pm$ 2.00} \\
4  & 44.30 $\pm$ 4.10 & \underline{45.90 $\pm$ 7.19} & 45.46 $\pm$ 5.32 & \textbf{46.63 $\pm$ 2.79} \\
5  & 59.09 $\pm$ 4.99 & 63.24 $\pm$ 4.16 & \underline{64.44 $\pm$ 4.82} & \textbf{64.81 $\pm$ 4.73} \\
6  & 15.69 $\pm$ 4.48 & 17.34 $\pm$ 5.53 & \textbf{18.07 $\pm$ 4.36} & \underline{17.73 $\pm$ 4.38} \\
7  & 30.02 $\pm$ 8.98 & 31.89 $\pm$ 5.97 & \underline{33.11 $\pm$ 7.40} & \textbf{36.99 $\pm$ 2.52} \\
8  & 11.82 $\pm$ 2.94 & 14.91 $\pm$ 6.09 & \underline{17.85 $\pm$ 8.51} & \textbf{19.47 $\pm$ 8.81} \\
9  & 55.94 $\pm$ 4.21 & 58.40 $\pm$ 2.32 & \underline{59.16 $\pm$ 3.07} & \textbf{62.58 $\pm$ 2.33} \\
10 & 13.57 $\pm$ 3.09 & \underline{13.80 $\pm$ 3.42} & 13.14 $\pm$ 1.97 & \textbf{15.48 $\pm$ 2.86} \\
11 & 26.85 $\pm$ 12.07& 28.43 $\pm$ 13.15& \textbf{38.17 $\pm$ 6.98} & \underline{35.93 $\pm$ 7.59} \\
12 & 8.14 $\pm$ 3.59  & 11.60 $\pm$ 4.57 & \textbf{13.54 $\pm$ 5.65} & \underline{13.48 $\pm$ 3.29} \\
\bottomrule
Avg &  32.76           &    35.36   &     \underline{36.49}     &     \textbf{38.10}   \\
\bottomrule
\end{tabular}
\end{table}

\paragraph{Observation.} The experimental results, as shown in Table~\ref{tab:heads}, demonstrate a gradual improvement in overall model performance with an increasing number of attention heads. This indirectly supports our earlier hypothesis regarding the relationship between the number of attention heads and the learning patterns of channel importance—more attention heads enable the model to capture more complex task-relevant inter-channel dependencies. Furthermore, since in our implementation the Transformer embedding dimension is defined as the product of the number of attention heads and the per-head embedding dimension, the possible values and upper bound of the number of heads are inherently constrained. By increasing the per-head embedding dimension, it is theoretically possible to explore more complex model configurations and further improve performance to some extent.

\subsection{Number of CATs}

\paragraph{Settings.} We set the number of CATs to 1, 4, 8, 12, and 16, while keeping the other network parameters fixed for the study.

\begin{table}[htbp]
\centering
\caption{Performance of different network parameter settings across subjects (best in \textbf{bold}, second best \underline{underlined}).}
\label{tab:CATs}
\begin{tabular}{cccccc}
\toprule
\multirow{2}{*}{subject} & \multicolumn{5}{c}{Accuracy $\pm$ Std} \\
\cmidrule(lr){2-6}
                         & 1       & 4       & 8       & 12       & 16       \\
\midrule
1  & 22.98 $\pm$ 10.09 & 53.67 $\pm$ 6.44  & \underline{60.77 $\pm$ 4.44}  & 60.15 $\pm$ 6.21  & \textbf{63.45 $\pm$ 4.29}  \\
2  & 39.24 $\pm$ 17.01 & 56.35 $\pm$ 4.30  & 58.62 $\pm$ 5.77  & \textbf{61.37 $\pm$ 4.43}  & \underline{60.09 $\pm$ 6.29}  \\
3  & 4.25 $\pm$ 2.14   & 7.54 $\pm$ 3.37   & \underline{8.89 $\pm$ 2.96}   & \textbf{9.95 $\pm$ 3.49}   & 7.44 $\pm$ 1.59   \\
4  & 15.81 $\pm$ 4.19  & 43.90 $\pm$ 4.77  & 44.30 $\pm$ 3.96  & \textbf{46.41 $\pm$ 3.80}  & \underline{45.83 $\pm$ 4.67}  \\
5  & 18.25 $\pm$ 3.25  & 53.19 $\pm$ 11.29 & 58.98 $\pm$ 5.21  & \underline{64.01 $\pm$ 3.14}  & \textbf{64.19 $\pm$ 3.76}  \\
6  & 8.90 $\pm$ 5.15   & 14.05 $\pm$ 4.69  & 16.42 $\pm$ 4.68  & \textbf{20.61 $\pm$ 3.87}  & \underline{18.79 $\pm$ 6.65}  \\
7  & 12.83 $\pm$ 6.33  & 21.66 $\pm$ 9.68  & \underline{29.67 $\pm$ 8.98}  & \textbf{31.52 $\pm$ 6.83}  & 26.26 $\pm$ 11.02 \\
8  & 3.92 $\pm$ 5.06   & 10.32 $\pm$ 6.99  & 12.40 $\pm$ 3.50  & \underline{16.16 $\pm$ 8.55}  & \textbf{16.90 $\pm$ 5.26}  \\
9  & 30.62 $\pm$ 9.97  & 51.50 $\pm$ 4.45  & \underline{56.16 $\pm$ 4.04}  & 56.07 $\pm$ 4.24  & \textbf{60.70 $\pm$ 2.87}  \\
10 & 3.42 $\pm$ 2.65   & 11.81 $\pm$ 2.48  & \textbf{13.39 $\pm$ 2.95}  & \underline{13.20 $\pm$ 4.27}  & 12.25 $\pm$ 2.48  \\
11 & 10.06 $\pm$ 6.76  & 26.85 $\pm$ 8.81  & 27.30 $\pm$ 13.28 & \underline{33.93 $\pm$ 8.95}  & \textbf{38.03 $\pm$ 15.15} \\
12 & 6.44 $\pm$ 4.72   & 9.29 $\pm$ 3.40   & 8.50 $\pm$ 3.59   & \underline{9.96 $\pm$ 6.29}   & \textbf{10.26 $\pm$ 4.41}  \\
\bottomrule
Avg & 14.73 & 30.01 & 32.95 & \underline{35.28} & \textbf{35.35} \\
\bottomrule
\end{tabular}
\end{table}

\paragraph{Observation.} As shown in Table~\ref{tab:CATs}, the overall model performance improves with an increasing number of CATs. This indirectly supports the argument made in the main text that a larger number of CATs can help the model capture more task-relevant channel importance patterns. However, the performance gain from increasing the number of CATs from 12 to 16 is marginal, suggesting that the representational capacity of CATs is limited when the amount of training data is fixed. An excessive number of CATs may even hinder model fitting. In addition, we observe considerable performance variation across subjects under the same CAT configuration, indicating that the difficulty of learning task-relevant features differs among individuals.

\subsection{Dropout Probability in the Transformer}

\paragraph{Settings.} We set the dropout probability in the Transformer to 0, 0.1, 0.2, 0.3, and 0.4, while keeping the other network parameters fixed for the study.

\begin{table}[htbp]
\centering
\caption{Performance of different network parameter settings across subjects (best in \textbf{bold}, second best \underline{underlined}).}
\label{tab:dropout}
\begin{tabular}{c c c c c c c}
\toprule
\multirow{2}{*}{Subject} & \multicolumn{5}{c}{Accuracy $\pm$ Std (\%)} \\
\cmidrule(lr){2-6}
 & 0 & 0.1 & 0.2 & 0.3 & 0.4 \\
\midrule
1  & 59.86 $\pm$ 5.11 & \textbf{60.44 $\pm$ 4.80} & 58.34 $\pm$ 3.60 & \underline{60.19 $\pm$ 4.46} & 58.92 $\pm$ 2.00 \\
2  & 57.33 $\pm$ 4.83 & 56.57 $\pm$ 4.66 & 59.41 $\pm$ 4.24 & \textbf{59.72 $\pm$ 5.15} & \underline{59.41 $\pm$ 2.80} \\
3  & \underline{10.18 $\pm$ 3.37} & 9.47 $\pm$ 3.49  & \textbf{10.27 $\pm$ 2.93} & 8.66 $\pm$ 2.69  & 8.89 $\pm$ 3.10 \\
4  & 38.36 $\pm$ 5.39 & 40.29 $\pm$ 6.57 & 42.62 $\pm$ 4.31 & \underline{43.53 $\pm$ 3.96} & \textbf{48.20 $\pm$ 4.34} \\
5  & 56.65 $\pm$ 4.66 & \textbf{59.67 $\pm$ 3.00} & \underline{59.53 $\pm$ 6.09} & 58.94 $\pm$ 5.16 & 55.92 $\pm$ 6.60 \\
6  & 15.33 $\pm$ 3.87 & 16.78 $\pm$ 5.48 & \textbf{17.41 $\pm$ 4.56} & 15.69 $\pm$ 4.48 & \underline{17.28 $\pm$ 5.93} \\
7  & 28.70 $\pm$ 10.43& 30.02 $\pm$ 9.47 & \textbf{32.08 $\pm$ 8.56} & 29.73 $\pm$ 9.00 & \underline{30.61 $\pm$ 6.04} \\
8  & 12.12 $\pm$ 3.80 & \underline{12.73 $\pm$ 5.01} & \textbf{13.07 $\pm$ 4.45} & 12.00 $\pm$ 3.08 & 9.43 $\pm$ 3.70 \\
9  & 55.49 $\pm$ 3.23 & 55.59 $\pm$ 4.48 & \textbf{58.52 $\pm$ 2.35} & 56.19 $\pm$ 4.03 & \underline{57.47 $\pm$ 4.07} \\
10 & 13.42 $\pm$ 3.74 & \textbf{14.85 $\pm$ 3.86} & \underline{14.09 $\pm$ 3.06} & 13.45 $\pm$ 2.99 & 12.03 $\pm$ 3.33 \\
11 & 22.71 $\pm$ 11.28& 27.23 $\pm$ 14.90& \textbf{27.90 $\pm$ 14.23}& \underline{27.34 $\pm$ 12.75}& 25.52 $\pm$ 14.76 \\
12 & \textbf{9.05 $\pm$ 2.35}  & 7.71 $\pm$ 1.82  & 8.38 $\pm$ 3.27  & \underline{8.80 $\pm$ 3.73}  & 7.04 $\pm$ 2.64 \\
\bottomrule
Avg & 31.60 & 32.61 & \textbf{33.47} & \underline{32.86} & 32.56 \\
\bottomrule
\end{tabular}
\end{table}

\paragraph{Observation.} As shown in Table~\ref{tab:dropout}, the optimal dropout probability appears to be around 0.2 overall. The primary purpose of dropout is to improve the model’s generalization ability; however, an excessively high dropout rate can hinder the model’s capacity to fit the data. Similar to the observations made with varying numbers of CATs, we also find substantial variation in performance across subjects under the same dropout setting. This further suggests that the quality and complexity of the data vary among different subjects.

\section{Important Channel Analysis with Du-IN Encoder}

In the main text, we employed EEGConformer as the downstream classifier to investigate the ability of EEGChaT in identifying task-relevant SEEG channels, and compared the selected channels with those manually selected. Theoretically, a similar purpose can also be achieved using the Du-IN Encoder model. Moreover, we expect that the superior classification capability of Du-IN Encoder can further enhance the channel selection performance of EEGChaT. A comparison between the classification performance using channels selected by EEGChaT with Du-IN Encoder as the downstream classifier and that using manually selected channels is presented in Table~\ref{tab:sel_chans_DuIN}. For clarity and aesthetics, \textbf{EC} refers to the channels selected by EEGChaT, while \textbf{MC} denotes the manually selected channels. All results were obtained based on the Du-IN Encoder model.

\begin{table}[htbp]
\centering
\caption{Performance of different methods across subjects (best in \textbf{bold}, second best \underline{underlined}). The Jaccard Index column is computed based on the overlap between the 10 hand-crafted channels and the 10 highest-weighted channels selected by EEGChaT (obtained by averaging the weights across all samples, attention heads, CATs, and random seeds).}
\label{tab:sel_chans_DuIN}
\resizebox{1\textwidth}{!}{
\begin{tabular}{cccccc}
\toprule
\multirow{2}{*}{Subject} & \multicolumn{4}{c}{Accuracy $\pm$ StdDev (\%)} & \multirow{2}{*}{\makecell{Jaccard\\ Index}} \\
\cmidrule(lr){2-5}
& MC w/ EEGChaT & MC w/o EEGChaT & EC w/ EEGChaT & EC w/o EEGChaT & \\
\midrule
1  & \underline{68.15 $\pm$ 4.10} & \textbf{73.18 $\pm$ 2.54} & 67.97 $\pm$ 1.92 & 66.45 $\pm$ 2.65 & 0.5385 \\
2  & 70.86 $\pm$ 3.13 & \underline{74.26 $\pm$ 1.56} & 71.17 $\pm$ 2.02 & \textbf{74.35 $\pm$ 2.35} & 0.6667 \\
3  & \underline{19.07 $\pm$ 3.52} & \textbf{28.15 $\pm$ 2.89} & 17.07 $\pm$ 2.79 & 17.10 $\pm$ 2.47 & 0.4286 \\
4  & 54.39 $\pm$ 4.12 & 54.64 $\pm$ 2.99 & \textbf{56.03 $\pm$ 4.24} & \underline{55.96 $\pm$ 3.57} & 0.8182 \\
5  & 74.97 $\pm$ 3.62 & \textbf{76.50 $\pm$ 2.26} & 74.97 $\pm$ 2.58 & \underline{75.37 $\pm$ 1.64} & 0.6667 \\
6  & \textbf{35.81 $\pm$ 3.69} & \underline{35.25 $\pm$ 3.29} & 33.17 $\pm$ 2.72 & 33.50 $\pm$ 1.84 & 0.5385 \\
7  & 45.07 $\pm$ 4.54 & \underline{46.35 $\pm$ 3.16} & \textbf{46.73 $\pm$ 1.78} & 41.47 $\pm$ 2.27 & 0.5385 \\
8  & \underline{45.33 $\pm$ 4.91} & \textbf{49.83 $\pm$ 2.31} & 26.48 $\pm$ 4.65 & 27.46 $\pm$ 4.01 & 0.4286 \\
9  & 64.94 $\pm$ 1.77 & \underline{69.16 $\pm$ 1.99} & 66.32 $\pm$ 2.36 & \textbf{70.53 $\pm$ 2.81} & 0.5385 \\
10 & \underline{18.08 $\pm$ 3.84} & \textbf{21.87 $\pm$ 1.74} & 15.35 $\pm$ 2.25 & 13.11 $\pm$ 1.58 & 0.4286 \\
11 & \textbf{66.67 $\pm$ 3.28} & 65.55 $\pm$ 1.90 & 61.97 $\pm$ 5.34 & \underline{65.62 $\pm$ 3.19} & 0.8182 \\
12 & \underline{41.89 $\pm$ 4.80} & \textbf{47.42 $\pm$ 1.01} & 25.08 $\pm$ 5.59 & 25.14 $\pm$ 2.25 & 0.4286 \\
\bottomrule
Avg & \underline{50.44} & \textbf{53.51} & 46.86 & 47.17 & 0.5698 \\
\bottomrule
\end{tabular}
}
\end{table}

The results indicate that when the downstream classifier is sufficiently powerful, EEGChaT struggles to consistently improve overall classification performance on suboptimal channel subsets, and may even have adverse effects. This could be attributed to suboptimal parameter settings of EEGChaT.

In addition, we applied the channels selected by EEGChaT with the assistance of the Du-IN Encoder to the EEGConformer model. The results are presented in Table~\ref{tab:sel_chans_DuIN_CFMR}.

\begin{table}[htbp]
\centering
\caption{Performance of different methods across subjects (best in \textbf{bold}).}
\label{tab:sel_chans_DuIN_CFMR}
\begin{tabular}{ccc}
\toprule
\multirow{2}{*}{Subject} & \multicolumn{2}{c}{Accuracy $\pm$ StdDev (\%)} \\
\cmidrule(lr){2-3}
 & EEGChaT+CFMR & CFMR \\
\midrule
1  & \textbf{61.71 $\pm$ 1.98} & 53.71 $\pm$ 1.89 \\
2  & \textbf{65.14 $\pm$ 2.28} & 61.83 $\pm$ 2.33 \\
3  & \textbf{14.20 $\pm$ 2.25} & 12.37 $\pm$ 1.20 \\
4  & \textbf{51.73 $\pm$ 3.40} & 45.97 $\pm$ 3.34 \\
5  & \textbf{67.76 $\pm$ 2.70} & 60.66 $\pm$ 3.66 \\
6  & 23.41 $\pm$ 1.84 & \textbf{23.87 $\pm$ 1.62} \\
7  & \textbf{40.47 $\pm$ 2.11} & 32.96 $\pm$ 1.92 \\
8  & \textbf{24.92 $\pm$ 2.60} & 20.14 $\pm$ 1.54 \\
9  & \textbf{58.21 $\pm$ 2.43} & 48.95 $\pm$ 2.59 \\
10 & \textbf{14.56 $\pm$ 1.63} & 11.43 $\pm$ 1.68 \\
11 & \textbf{55.66 $\pm$ 2.87} & 47.95 $\pm$ 2.91 \\
12 & \textbf{16.51 $\pm$ 3.52} & 19.25 $\pm$ 2.35 \\
\bottomrule
Avg & \textbf{41.19} & 36.59 \\
\bottomrule
\end{tabular}
\end{table}

The results indicate that, compared to the channels selected by EEGChaT using EEGConformer as the classifier, those selected when using Du-IN Encoder exhibit stronger task relevance. The classification performance improved by 2.01\% and 2.64\% compared to the results of 39.18\% and 33.95\% reported in Table~\ref{tab:sel_chans_EEGConformer}, respectively. These results suggest that employing a more powerful classifier in the SEEG channel selection task can enhance the channel selection capability of EEGChaT.

\section{Ablation Study}

In this section, we validate the necessity of the proposed mechanisms in the main methodology using the EEGConformer classifier.

\subsection{Impact of Attention Rollout}

In this subsection, we investigate the importance of the Attention Rollout mechanism for channel selection in EEGChaT. Three conditions are considered: using the improved Attention Rollout (\textbf{ARO-H}, which preserves the attention head dimension), using the original Attention Rollout (\textbf{ARO-Avg}, which averages over attention heads), and not using Attention Rollout at all (\textbf{No-ARO}). In the \textbf{No-ARO} setting, the attention matrix from the last Transformer layer is directly used in place of the Attention Rollout. The results are presented in Table~\ref{tab:rollout}.

\begin{table}[htbp]
\centering
\caption{Performance of different methods across subjects (best in \textbf{bold}, second best \underline{underlined}).}
\label{tab:rollout}
\begin{tabular}{cccc}
\toprule
\multirow{2}{*}{Subject} & \multicolumn{3}{c}{Accuracy ± StdDev (\%)} \\
\cmidrule(lr){2-4}
 & ARO-H & ARO-Avg & No-ARO \\
\midrule
1  & \textbf{60.26 ± 4.44} & \underline{57.15 ± 3.70} & 52.70 ± 4.43 \\
2  & \textbf{58.62 ± 5.77} & \underline{55.10 ± 10.72} & 46.80 ± 13.38 \\
3  & \textbf{8.86 ± 2.89} & \underline{8.63 ± 4.46} & 8.34 ± 1.25 \\
4  & \underline{44.59 ± 4.29} & 40.44 ± 6.32 & \textbf{51.26 ± 2.63} \\
5  & \underline{58.94 ± 5.16} & 55.88 ± 8.86 & \textbf{59.45 ± 4.38} \\
6  & \textbf{15.69 ± 4.48} & 13.98 ± 4.20 & \underline{14.31 ± 6.60} \\
7  & \textbf{30.02 ± 8.98} & 26.23 ± 9.05 & \underline{27.86 ± 6.02} \\
8  & \textbf{11.85 ± 2.97} & 8.02 ± 2.60 & \underline{8.33 ± 5.22} \\
9  & \textbf{56.19 ± 4.03} & \underline{53.35 ± 6.05} & 52.78 ± 5.52 \\
10 & \textbf{13.39 ± 2.95} & \underline{10.41 ± 3.46} & 8.83 ± 3.00 \\
11 & \textbf{28.46 ± 11.92} & \underline{26.64 ± 12.27} & 18.19 ± 10.24 \\
12 & \textbf{8.80 ± 3.73} & 8.56 ± 2.62 & \underline{6.19 ± 2.27} \\
\bottomrule
Avg & \textbf{32.82} & 29.66 & \underline{30.44} \\
\bottomrule
\end{tabular}
\end{table}

The results demonstrate that our approach effectively enhances the channel selection capability of EEGChaT (\textbf{ARO-H}). The performance drop observed in the \textbf{ARO-Avg} setting highlights the importance of preserving the attention head dimension, indicating that different attention heads are capable of learning distinct channel combinations, which are then leveraged by the downstream classifier. Moreover, directly using the attention matrix from the last Transformer layer as a substitute for Attention Rollout (\textbf{No-ARO}) also enables a certain degree of channel importance analysis. This may be attributed to the relatively shallow architecture (two Transformer layers), where the model tends to concentrate the learned inter-channel dependencies in the final attention layer.

\subsection{Impact of Attention Masking}

\begin{figure}[htbp] 
  \centering
  \includegraphics[width=\linewidth]{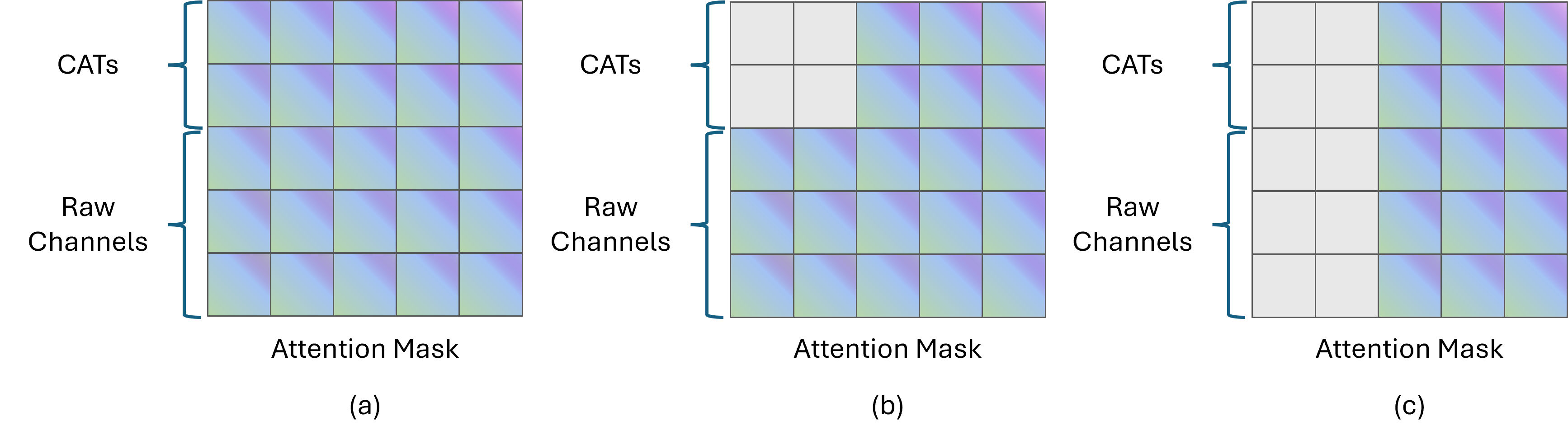}
  \caption{Different attention matrix masking strategies.}
  \label{fig:AttentionMask}
\end{figure}

In this subsection, we investigate different attention matrix masking strategies. An effective masking strategy can help CATs aggregate information from different channels without mutual interference. We consider three strategies: no attention masking (as shown in Fig.~\ref{fig:AttentionMask}a), the masking strategy illustrated in Fig.~\ref{fig:AttentionMask}b, and the one shown in Fig.~\ref{fig:AttentionMask}c. The corresponding results are presented in Table~\ref{tab:attentionMask}.

\begin{table}[htbp]
\centering
\caption{Performance of different methods across subjects (best in \textbf{bold}, second best \underline{underlined}).}
\label{tab:attentionMask}
\begin{tabular}{cccc}
\toprule
\multirow{2}{*}{Subject} & \multicolumn{3}{c}{Accuracy $\pm$ StdDev (\%)} \\
\cmidrule(lr){2-4}
 & a & b & c \\
\midrule
1  & 59.65 $\pm$ 3.60 & \underline{60.12 $\pm$ 3.74} & \textbf{60.26 $\pm$ 4.44} \\
2  & \underline{58.86 $\pm$ 5.21} & \textbf{59.47 $\pm$ 4.22} & 58.62 $\pm$ 5.77 \\
3  & \textbf{9.95 $\pm$ 2.25} & \underline{9.37 $\pm$ 2.85} & 8.86 $\pm$ 2.89 \\
4  & \underline{43.72 $\pm$ 4.75} & 42.59 $\pm$ 4.56 & \textbf{44.59 $\pm$ 4.29} \\
5  & 56.03 $\pm$ 5.64 & \underline{56.98 $\pm$ 6.60} & \textbf{58.94 $\pm$ 5.16} \\
6  & \underline{16.02 $\pm$ 3.73} & \textbf{16.22 $\pm$ 4.21} & 15.69 $\pm$ 4.48 \\
7  & \underline{30.30} $\pm$ 8.33 & \textbf{30.86 $\pm$ 9.21} & 30.02 $\pm$ 8.98 \\
8  & \underline{12.03 $\pm$ 5.18} & \textbf{12.43 $\pm$ 4.26} & 11.85 $\pm$ 2.97 \\
9  & 55.52 $\pm$ 4.38 & \textbf{56.42 $\pm$ 4.17} & \underline{56.19 $\pm$ 4.03} \\
10 & \underline{13.54 $\pm$ 2.62} & \textbf{13.68 $\pm$ 2.42} & 13.39 $\pm$ 2.95 \\
11 & \textbf{28.01 $\pm$ 14.90} & 25.73 $\pm$ 11.93 & \underline{26.64 $\pm$ 12.27} \\
12 & \underline{8.44 $\pm$ 4.17} & 7.65 $\pm$ 3.01 & \textbf{8.80 $\pm$ 3.73} \\
\bottomrule
Avg & 32.59 & \underline{32.63} & \textbf{32.82} \\
\bottomrule
\end{tabular}
\end{table}

The results suggest that varying masking strategies exert only a limited effect on the final outcomes, indicating that with adequate data, EEGChaT is capable of exploiting the Transformer’s intrinsic ability to model inter-channel dependencies, without requiring explicit mechanisms for information propagation.

\section{Cross-subject Experiments}

\begin{figure}[htbp] 
  \centering
  \includegraphics[width=\linewidth]{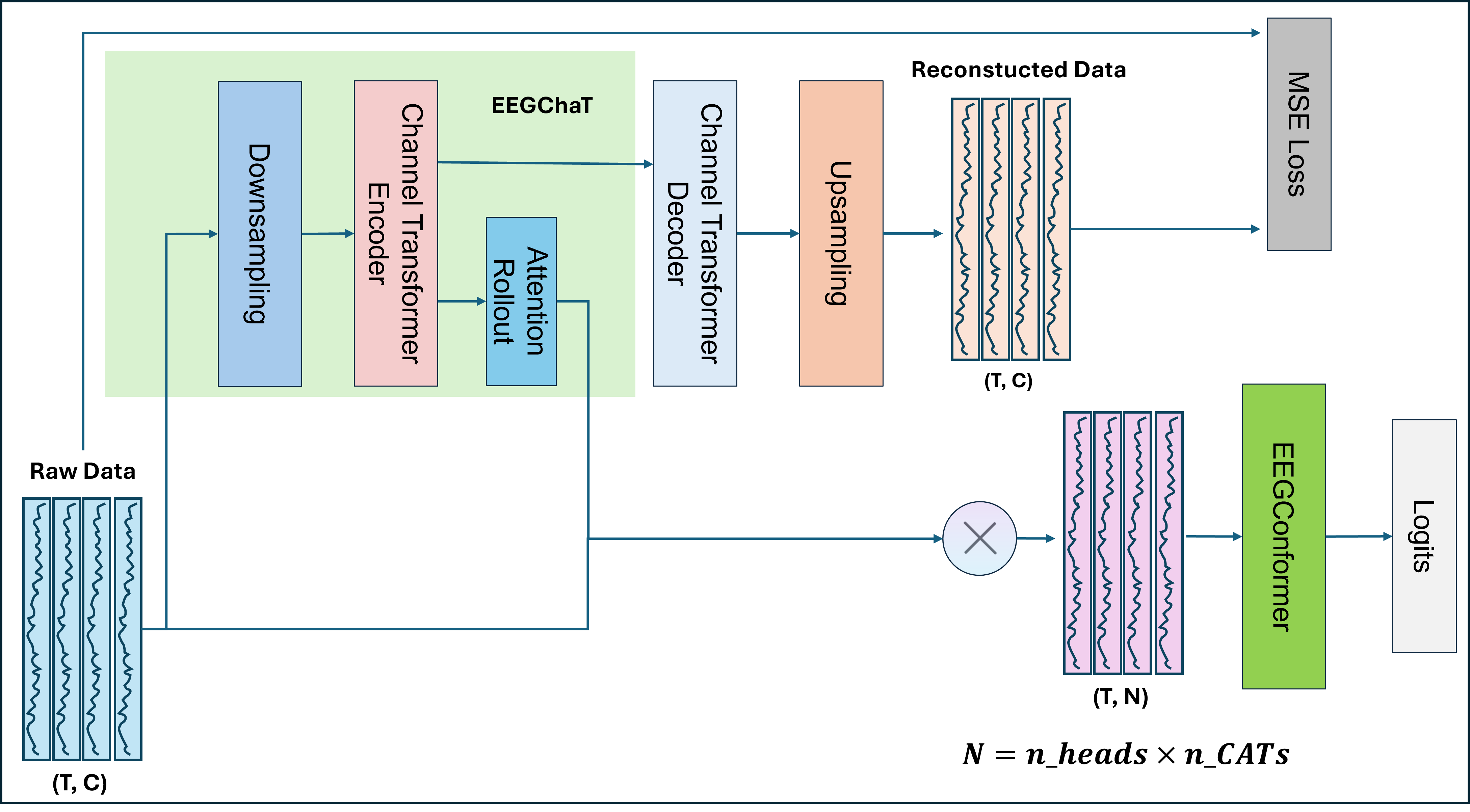}
  \caption{The overall training pipeline combining cross-subject pretraining and downstream task fine-tuning.}
  \label{fig:pipeline}
\end{figure}

Since the channel dimension of SEEG signals is treated as a token sequence, our method not only possesses the ability to perform channel selection, but also supports cross-subject data training without requiring additional alignment methods. We design the training pipeline as illustrated in Fig.~\ref{fig:pipeline}, where the EEGChaT module utilizes an autoencoder to learn inter-channel dependencies from unlabeled data.

After pretraining, the learned EEGChaT weights are fine-tuned on downstream classification tasks to enhance channel selection performance. Due to the scarcity of high-quality unlabeled SEEG data, we partition each subject's data in the DuIN dataset into four subsets with a $1:7:1:1$ ratio for pretraining, training, validation, and testing, respectively. To facilitate data loading, the channel dimensions across all subjects’ training data are padded to a consistent size.

The EEGChaT architecture adopts the optimal hyperparameters identified in Appendix~\ref{sec:appendix_A}:
\begin{itemize}
    \item Number of \textbf{Transformer layers}: 2
    \item Number of \textbf{attention heads}: 10
    \item Number of \textbf{CATs}: 16
    \item \textbf{Dropout probability} in the Transformer: 0.2
\end{itemize}
These configurations are chosen to improve EEGChaT’s capacity for channel selection, thereby enabling robust analysis of channel importance in complex cross-subject scenarios.

The downstream classifier adopts the Du-IN Encoder. To evaluate the impact of pretraining on the performance of EEGChaT, we also conducted experiments under the same data split without EEGChaT pretraining. The experimental results are summarized in Table~\ref{tab:cross}.

\begin{table}[htbp]
\centering
\caption{Performance of different methods across subjects (best in \textbf{bold}).}
\label{tab:cross}
\begin{tabular}{ccc}
\toprule
\multirow{2}{*}{Subject} & \multicolumn{2}{c}{Accuracy $\pm$ StdDev (\%)} \\
\cmidrule{2-3}
 & w/ pretraining & w/o pretraining \\
\midrule
1 & 66.12 $\pm$ 3.22 & \textbf{66.30} $\pm$ 3.00 \\
2 & 71.69 $\pm$ 2.47 & \textbf{72.33} $\pm$ 2.10 \\
3 & \textbf{6.54} $\pm$ 1.95 & 5.31 $\pm$ 0.65 \\
4 & 41.28 $\pm$ 4.18 & \textbf{51.07} $\pm$ 4.51 \\
5 & 66.12 $\pm$ 3.15 & \textbf{71.51} $\pm$ 3.03 \\
6 & 13.06 $\pm$ 4.20 & \textbf{19.62} $\pm$ 1.84 \\
7 & 30.49 $\pm$ 5.24 & \textbf{41.41} $\pm$ 3.53 \\
8 & 13.35 $\pm$ 3.03 & \textbf{15.03} $\pm$ 4.27 \\
9 & 58.30 $\pm$ 3.64 & \textbf{66.89} $\pm$ 0.97 \\
10 & \textbf{10.35} $\pm$ 1.41 & 8.01 $\pm$ 2.48 \\
11 & 29.93 $\pm$ 10.70 & \textbf{35.72} $\pm$ 8.27 \\
12 & 12.81 $\pm$ 4.65 & \textbf{14.27} $\pm$ 5.12 \\
\bottomrule
Avg & 35.00 & \textbf{38.96}  \\
\bottomrule
\end{tabular}
\end{table}

Although EEGChaT was configured with more suitable hyperparameters, its overall performance was inferior to the results presented in the main text. This degradation is primarily attributed to the reduced size of the training set. The observation highlights that SEEG decoding tasks demand a large amount of high-quality data to achieve optimal performance.

Moreover, the average performance of pretrained EEGChaT across all subjects was lower than that of the non-pretrained counterpart. This may be because the pretraining process was not aligned with the objectives of the downstream classification task. During the reconstruction phase using the autoencoder, the model may have focused on learning channel dependencies that reduce reconstruction loss rather than those that are truly relevant to the classification objective.

Finally, due to the limited amount of data, the number of samples available for pretraining was insufficient. We trained the model for 500 epochs on approximately 4,000 samples, achieving a final mean squared error (MSE) loss of around 42. To fully realize the potential of EEGChaT for cross-subject training, a substantially larger dataset and a more task-aligned pretraining strategy are necessary.

\section{Computational Resources}

All experiments were conducted on a computing cluster equipped with the following specifications:

\begin{itemize}
  \item \textbf{GPU:} NVIDIA V100 32GB
  \item \textbf{Memory:} 256 GB RAM
  \item \textbf{Operating System:} Ubuntu 22.04 LTS
  \item \textbf{Framework:} PyTorch 2.3.1 with CUDA 12.2
\end{itemize}

The pretraining of EEGChaT takes approximately \textbf{2.19 hours} (500 epochs, batch size 160). The training time for a single classification task depends on the network architecture, with the longest case taking about \textbf{0.56 hours} (200 epochs, batch size 32).

\end{document}